\documentclass[10pt,twocolumn,letterpaper]{article}

\usepackage[dvipsnames,svgnames,table]{xcolor} 

\usepackage[pagenumbers]{cvpr} 

\usepackage[dvipsnames]{xcolor}
\newcommand{\cm}{{\ding{51}}}
\newcommand{\xm}{{\ding{55}}}

\usepackage{times}
\usepackage{epsfig}
\usepackage{graphicx}
\usepackage{amsmath}
\usepackage{amssymb}
\usepackage{bm}
\usepackage{graphbox}

\usepackage{makecell}
\usepackage{booktabs}
\usepackage{tabularx}
\usepackage{multirow}
\usepackage{pifont}
\usepackage{tablefootnote}
\usepackage[para,online,flushleft]{threeparttable}

\usepackage{algorithm}
\usepackage{setspace}
\usepackage[noend]{algpseudocode}

\usepackage{color}

\definecolor{cvprblue}{rgb}{0.21,0.49,0.74}
\usepackage[pagebackref,breaklinks,colorlinks,allcolors=cvprblue]{hyperref}

\def\paperID{*****} 
\def\confName{CVPR}
\def\confYear{2026}

\title{SneakPeek: Future-Guided Instructional Streaming Video Generation}
\def\Name{SneakPeek}

\author{
Cheeun Hong$^\dagger$$^\ddagger$\thanks{Work done during internship at Meta Superintelligence Labs. $^\dagger$Meta Superintelligence Labs. $^\ddagger$Seoul National University. Correspondence to: \tt{cheeun914@snu.ac.kr}}
\and
German Barquero$^\dagger$
\and
Fadime Sener$^\dagger$
\and
Markos Georgopoulos$^\dagger$
\and
Edgar Schönfeld$^\dagger$
\and
Stefan Popov$^\dagger$
\and
Yuming Du$^\dagger$
\and
Oscar Mañas$^\dagger$
\and
Albert Pumarola$^\dagger$
}

\begin{document}
\maketitle

\begin{abstract}
Instructional video generation is an emerging task that aims to synthesize coherent demonstrations of procedural activities from textual descriptions. 
Such capability has broad implications for content creation, education, and human-AI interaction, yet existing video diffusion models struggle to maintain temporal consistency and controllability across long sequences of multiple action steps.
We introduce a pipeline for future-driven streaming instructional video generation, dubbed \textbf{\Name}, a diffusion-based autoregressive framework designed to generate precise, stepwise instructional videos conditioned on an initial image and structured textual prompts.
Our approach introduces three key innovations to enhance consistency and controllability: 
(1) predictive causal adaptation, where a causal model learns to perform next-frame prediction and anticipate future keyframes;
(2) future-guided self-forcing with a dual-region KV caching scheme to address the exposure bias issue at inference time; 
(3) multi-prompt conditioning, which provides fine-grained and procedural control over multi-step instructions.
Together, these components mitigate temporal drift, preserve motion consistency, and enable interactive video generation where future prompt updates dynamically influence ongoing streaming video generation. 
Experimental results demonstrate that our method produces temporally coherent and semantically faithful instructional videos that accurately follow complex, multi-step task descriptions.
\end{abstract}
\section{Introduction}
\label{sec:intro}

Fine-grained generation of human-object interactions lies at the heart of modeling and simulating how humans act upon the world. Recent advances in video generation~\cite{liu2024sora,polyak2024movie,wan2025wan,seawead2025seaweed,agarwal2025cosmos} have greatly improved visual quality and realism, yet state-of-the-art~\cite{wan2025wan,seawead2025seaweed} models still struggle to capture fine-grained human-object interactions, often producing inaccurate contact and physically implausible motion.

Beyond fine-grained accuracy, generating long-horizon videos remains challenging despite progress in text-to-video generative models~\cite{yang2024cogvideox,chen2025skyreels,zhao2025riflex}. While diffusion-based models (DiTs)~\cite{wan2025wan,yang2024cogvideox,liu2024sora,polyak2024movie} can produce high-quality short clips, their reliance on bidirectional attention results in substantial computational cost. On the other hand, autoregressive causal models~\cite{yin2025slow,teng2025magi} enable more efficient inference by leveraging cached temporal states. However, maintaining temporal consistency and semantic controllability across sequential actions remain a fundamental challenge, as autoregressive diffusion models tend to accumulate prediction errors over time, leading to motion freezing or divergence from the intended action.

Such long-horizon fine-grained interactions are most naturally and abundantly found in instructional egocentric videos~\cite{grauman2024ego,damen2018scaling}, which capture people performing tasks from a first-person perspective over minute-long durations and remain highly challenging even for visual understanding~\cite{nagarajan2023egoenv,tan2023egodistill,shapovalov2023replay,zhang2025exo2ego,jia2022egotaskqa,xue2025progress,chatterjee2025streaming}. These videos also have broad applications in skill learning, robotic imitation, and education, where visual demonstrations effectively convey how to perform complex tasks. Recently, there has been growing interest in generating such videos~\cite{li2024handi,liu2024exocentric,li2024egogen}, yet existing works either focus solely on image generation~\cite{lu2024multimodal,lai2024lego,souvcek2025showhowto,suo2025long} or are limited to short durations~\cite{li2024handi}.

To address these challenges, we introduce \Name{}, a future-guided autoregressive DiT framework for streaming instructional videos that maintains strong temporal consistency and fine-grained semantic control over long horizons. \Name{} models the relationship between short- and long-term dynamics through three key components:

\begin{itemize}
    \item We introduce predictive causal adaptation, a method that transforms a bidirectional video generation model into a causal one capable of operating in two modes: next-frame prediction and future keyframe generation. In doing so, the causal model learns to capture both short- and long-term scene dynamics.

    \item We present future-guided Self-Forcing, which addresses one of the main limitations of Self-Forcing~\cite{huang2025self}: long-term drift. Specifically, we condition each next-frame prediction not only on past frames but also on a previously generated future keyframe. This future keyframe acts as an anchor that mitigates error accumulation and encourages the model to plan ahead, resulting in smoother, more coherent, and more realistic videos. 
    
    \item We apply and validate our streaming video model on a challenging benchmark: fine-grained procedural control. To this end, we extend classic prompt conditioning in video diffusion models to support two levels of prompting: a global procedural description and fine-grained action instructions. As a result, \Name{} can stream complex procedural videos with strong global consistency and fine-grained controllability. Our experiments show that \Name{} produces videos with higher temporal coherence and more accurate adherence to multi-step task descriptions than prior approaches.

\end{itemize}

\section{Related works}
\label{sec:related_works}

\subsection{Long video generation}
\label{subsec:long_video_generation}

Generating long and temporally coherent videos remains a fundamental challenge in video diffusion models due to accumulated prediction errors that degrade motion consistency and visual quality over time. DiT-based models~\cite{wan2025wan,yang2024cogvideox,liu2024sora,polyak2024movie,lu2023vdt,chen2024gentron,menapace2024snap,gao2024vid} can generate high-quality short clips, but struggle to scale to longer sequences. Their reliance on bidirectional attention prevents the use of key-value (KV) caching, since each frame attends to both past and future contexts. As a result, attention is required to be recomputed at every step, leading to substantial computational overhead. In contrast, autoregressive models~\cite{yin2025slow,teng2025magi,kodaira2025streamdit,yin2023nuwa,henschel2025streamingt2v,xie2025progressive} with causal attention enable more efficient inference by reusing cached KVs, but they often suffer from temporal drift and quality degradation as the sequence length increases. Recent works~\cite{chen2024diffusion,huang2025self,chen2025skyreels} explore various strategies to address this issue by introducing self-conditioning or self-feedback to reduce errors during autoregressive rollouts. Furthermore, for further efficient generation, a frame-packing strategy~\cite{zhang2025packing} that compresses multiple frames into a single latent representation is introduced. However, these methods lack fine-grained control over step-wise instructions, limiting their effectiveness to repetitive or low-motion tasks (\eg, walking). Concurrently,~\citet{zhang2025framecontextpackingdrift} extends FramePack~\cite{zhang2025packing} to support multi-prompt conditioning. While it retains long-range history through compressed tokens, it is not explicitly trained for predictive reasoning for future frames under missing temporal contexts. Meanwhile, our method learns how to predict far-future states from the bidirectional backbone, which are then used to condition the next frame generation.

\subsection{Instruction-based generation}
\label{subsec:instruction_based_generation}

Instruction-based visual generation has recently gained attention as a means to synthesize videos that depict step-by-step procedures aligned with natural language instructions. Along with the effort to build egocentric datasets~\cite{grauman2024ego,damen2018scaling,sener2022assembly101, zhukov2019cross} that capture multi-step instructions, existing approaches~\cite{lu2024multimodal,lai2024lego,souvcek2025showhowto,suo2025long} focus on keyframe-level generation for multi-step reasoning. However, these methods remain limited to discrete step visualization without dynamic continuity. Recent works~\cite{yu2025veggie,zhang2024effived,xing2023vidiff,qin2024instructvid2vid} extend to video domain, aiming to edit videos based on instructions. Yet, they are limitedly focused on local editing tasks like adding and removing, rather than producing motions to execute complex procedural tasks. Meanwhile, HANDI~\cite{li2024handi}, a hand-centric text-and-image conditioned video generation model, moves closer to true video synthesis but is constrained to short single-action sequences with localized hand motion. In contrast, our work addresses full-sequence instructional video generation, modeling both global and step-specific textual prompts to produce long-horizon, temporally consistent videos that visually execute complex procedural tasks.
\section{Proposed method}
\label{sec:proposed_method}

In this paper, our goal is to synthesize egocentric instructional videos for multi-step tasks from textual instructions in a streaming fashion. This task requires generating long, temporally coherent sequences that accurately follow fine-grained, step-level prompts describing distinct procedural actions. However, when existing video generation architectures are applied autoregressively, they often suffer from temporal drift, visual inconsistency, and a loss of precise step-level controllability over long horizons. To address these challenges, we introduce a future-guided autoregressive generation framework designed to ensure long-term coherence and fine-grained procedural control.

Our approach integrates three key components as a unified framework: \textbf{Predictive Causal Adaptation}, which predicts future keyframes under partial context (Section~\ref{subsec:predictive_causal_adaptation}); \textbf{Future-guided Self-Forcing}, which learns bidirectional temporal dependency (Section~\ref{subsec:future_based_self_forcing}); and \textbf{Temporal Masking} on cross-attention that provides fine-grained control across step-wise instruction prompts (Section~\ref{subsec:fine_grained_instruction_control}). The following sections describe each component in detail.

\subsection{Predictive causal adaptation}
\label{subsec:predictive_causal_adaptation}

\begin{figure}
    \centering
    \includegraphics[width=1\linewidth]{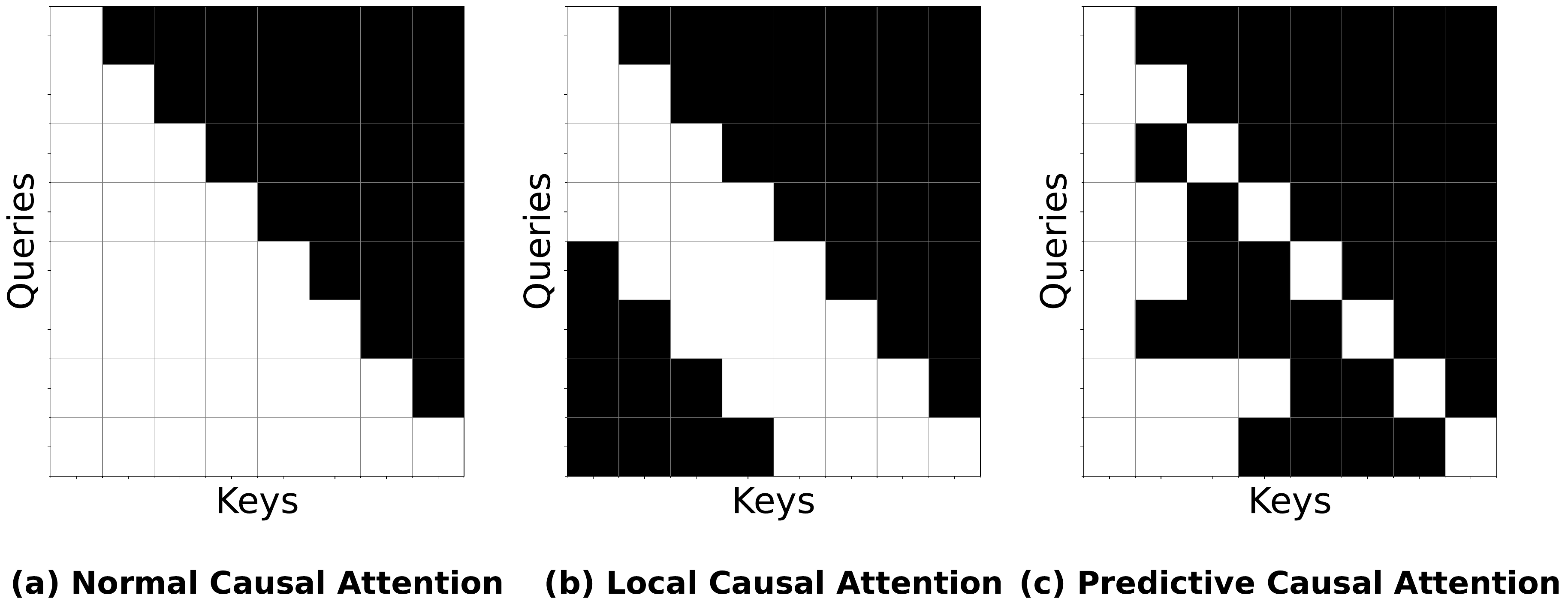}
    \caption{
        \textbf{Causal attention mask for predictive causal distillation.}  
        We use a random causal mask that selectively reveals (white) only a continuous segment of past frames while masking out the remaining history.
        This encourages the model to learn temporally coherent motion prediction even when intermediate contextual frames are unavailable (\ie, temporal jump).
    }
    \label{fig:method-mask}
\end{figure}

The first step toward a streaming video diffusion model is to obtain a generator capable of next-frame prediction, where each frame is produced conditioned solely on previously generated frames (i.e., operating in a fully causal manner). To achieve this, we adapt a pretrained bidirectional diffusion model by finetuning it with a causal attention mask (\Cref{fig:method-mask}a) and independent per-frame noise schedules, following the Diffusion-Forcing strategy~\cite{chen2024diffusion}. Through this process, the causalized model retains the broad visual and semantic priors of the original bidirectional backbone while enabling autoregressive frame generation. However, as discussed before, autoregressive generation in continuous domains such as images and videos is highly susceptible to exposure bias~\cite{schmidt2019generalization}, where small prediction errors accumulate over time. These issues are particularly damaging in instructional video generation, where actions must remain fine-grained, visually reliable, and precisely aligned with step-level instructions.

To address these limitations, we instead finetune the base model on two tasks: next-frame prediction and future keyframe (KF) prediction. This encourages the model to preserve the long-range reasoning and planning abilities inherited from the original bidirectional backbone; capabilities that are typically lost when switching to a purely autoregressive formulation. We enable both tasks with a simple change to the Transformer~\cite{peebles2023scalable}’s causal attention mask, without modifying the model architecture or training losses. For each generated frame (query), we randomly disable attention to a continuous span of preceding frames, see \Cref{fig:method-mask}c. If the blocked span is empty, the model learns next-frame prediction; if the span hides $N$ previous frames, the model is instead supervised to predict a KF $N$ steps into the future. As a result of this predictive causal adaptation (PCA), the model naturally supports both next-frame prediction and future KF forecasting at any temporal offset within the training horizon, and can operate with arbitrary amounts of context. However, it still inherits the exposure-bias issues of autoregressive generation, causing streaming video quality to drift over time. We address this limitation in the next section.

\subsection{Future-guided self-forcing}
\label{subsec:future_based_self_forcing}

After PCA, given a video sequence of $T$ frames, each frame $x_i (i{=}1,...,T)$ can be autoregressively predicted using the chain rule of conditional distribution as follows:
\begin{equation}
    p(x_{1:T})= \Pi^T_{i=1} p(x_i|x_{<i}),
    \label{eq:prob-base}
\end{equation}
where $x_{<i}$ denotes previously generated frames and each $p(x_i|x_{<i})$ is computed with a diffusion process that progressively denoises Gaussian noise conditioned on past frames. At this stage, our model trained with PCA can also predict future KFs at any temporal offset, allowing it to anticipate video dynamics several seconds into the future with a single autoregressive step. This means the model can generate KFs that span minutes of content using only a few autoregressive iterations, shifting the onset of exposure-bias-induced degradation from a few seconds to several minutes. We leverage this property to limit error accumulation during next-frame prediction by changing the model’s operating mode.

Instead of conditioning next-frame prediction solely on past frames, we prompt the model to generate a KF at a temporal horizon of $F$ frames and then condition the autoregressive next-frame generation on this KF. As a result, the model is guided toward a known future state, producing video that transitions smoothly toward that target, and its visual quality is tied to that of the KF. More formally, we reformulate the autoregressive chain as:
\begin{equation}
    p(x_{1:T})= \Pi^T_{i=1} p(x_i|x_{<i}, x_F),
    \label{eq:prob}
\end{equation}
where $x_F$ denotes a predicted future KF at the first iteration. Since $p(x_F)=p(x_F|x_1)$, exposure bias affects only one autoregressive step. When generating $x_{F+1}$, the model can attend to a high-quality frame that has accumulated almost no error. These predicted future KFs act as temporal anchors that stabilize long-term motion and reduce drift. This enables the model to bridge temporal gaps by generating plausible intermediate frames that connect the current state to a known future target.

\begin{figure}
    \centering
    \includegraphics[width=1\linewidth]{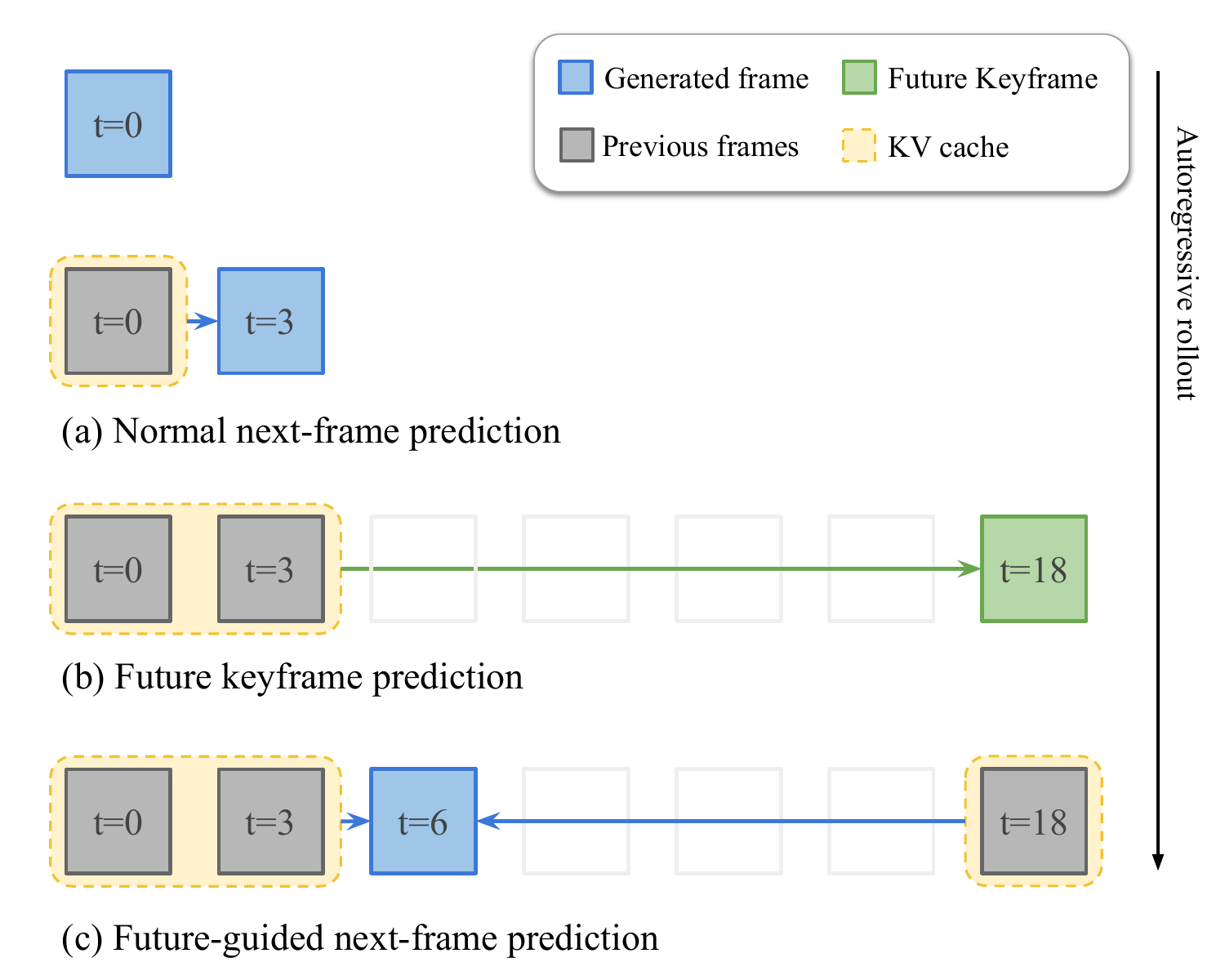}
    \caption{
        \textbf{Different sampling modes of our framework.}
        (a) Normal next-frame prediction.
        (b) Future KF prediction.
        (c) Future-guided next-frame prediction, where next frames are generated by attending to \textit{both} the previously generated sequences and the predicted future sequence, from the KV cache.
    }
    \label{fig:method-sampling}
\end{figure}

To support this new operating mode while reducing exposure bias, we introduce future-guided Self-Forcing. In this variant of Self-Forcing~\cite{huang2025self}, the next-frame prediction task is no longer trained solely on the model’s own generated frames; instead, the model also attends to a future KF that it previously produced, following Equation~\ref{eq:prob}. As in PCA, the model is trained on both tasks: (1) next-frame prediction and (2) future KF prediction.
As shown in \Cref{fig:method-sampling}, the next-frame mode generates frames autoregressively, conditioned on past frames (\Cref{fig:method-sampling}a). At any point, the model can be instructed to generate a future KF (\Cref{fig:method-sampling}b); from that moment on, subsequent next-frame predictions are conditioned on both the historical frames and the newly predicted future KF (\Cref{fig:method-sampling}c).
To optimize compute during training and inference, we employ a KV cache that stores the key and value projections from the attention layers of each fully denoised frame. To support both prediction modes, we introduce a dual-region KV cache that jointly stores keys and values for past frames and future KFs. These entries are stored with their original RoPE embeddings, ensuring that temporal distances for both historical frames and the future KF are correctly updated for every new query. When the cache reaches its maximum capacity, a sliding-window mechanism removes the oldest entries separately for the two cache regions (past frames and future KFs), ensuring efficient and stable conditioning over long sequences.
By randomizing the moment at which the model is prompted to generate a new KF, future-guided Self-Forcing makes the next-frame prediction task more robust to errors accumulated in both past frames and the predicted future KF.

\subsection{Temporal masking for fine-grained instruction}
\label{subsec:fine_grained_instruction_control}

Instructional and procedural videos often consist of multiple semantic segments, each corresponding to a distinct step within a larger task. For example, an instructional video on ``cooking a chicken'' may include stages such as ``washing the chicken'' followed by ``frying the chicken.'' Each step is semantically distinct and typically described by its own textual instruction, while the entire sequence aligns with a global task-level prompt. However, conventional cross-attention mechanisms in video diffusion models condition the entire sequence on a single global prompt, limiting the ability to enforce fine-grained temporal control and accurate step-wise alignment.

To address this, we add temporal masking for cross-attention, which enables temporally localized conditioning while maintaining access to the global context. Specifically, during generation, we construct temporal masks that restrict each query (corresponding to a latent frame) to attend only to text tokens associated with its corresponding temporal segment, while still allowing attention to the global instruction prompt. This ensures that each segment of the generated video adheres to its respective textual description, yet remains semantically consistent with the main goal. Moreover, we apply separate temporal masks for the future KF and the next-frame prediction tasks, enabling the model to predict the future KF conditioned on a new text prompt that describes the upcoming action, rather than reusing the ongoing text instruction. This allows the model to better anticipate diverse future motion and scene transitions.

\section{Experiments}
\label{sec:experiments}
\subsection{Data and metrics}

\begin{table*}[t]
\renewcommand{\arraystretch}{1.1}
\centering
\aboverulesep=0ex
\belowrulesep=0ex
\caption{ 
    \textbf{Comparison with prior long video generation methods}.
    For a fair evaluation on multi-prompt data, we apply cross attention masking for step-wise multi-prompt  control (MP) to baseline methods.
}
\resizebox{0.99\textwidth}{!}{
    \begin{tabular}{l|cc cc cc}
        \toprule
        \multirow{2}{*}{Method} & \multicolumn{2}{c}{Text-Video Consistency} & \multicolumn{2}{c}{Image-Video Consistency} & {Video Smoothness} & Video Similarity \\
        \cmidrule(lr){2-3} \cmidrule(lr){4-5} \cmidrule(lr){6-6} \cmidrule(lr){7-7} 
         & ViCLIP$_\uparrow$ & ViCLIP-M$_\uparrow$ & ImageCLIP$_\uparrow$ & ViCLIP-I$_\uparrow$ & CLIP-F$_\uparrow$ & EgoVLP$_\uparrow$ \\
        \midrule
        Baseline          & 0.1376 & 0.1292 & 0.6804 & 0.6659 & 0.9604 & 0.3478 \\
        + CausVid~\cite{yin2025slow}       & 0.1399 & 0.1332 & 0.6837 & 0.5533 & 0.9502 & 0.3540 \\
        + SelfForcing~\cite{huang2025self} & 0.1515 & 0.1498 & 0.7903 & 0.6966 & 0.9606 & 0.4987 \\
        + FramePack~\cite{zhang2025framecontextpackingdrift} & 0.1612 & 0.1558 & \textbf{0.8154} & 0.7694 & 0.9575 & 0.5369 \\
        + \Name (\textbf{Ours}) &  \textbf{0.1663} & \textbf{0.1609} & \underline{0.8153} & \textbf{0.7750} & \textbf{0.9722} & \textbf{0.5445} \\
        \bottomrule
    \end{tabular}
}
\label{tab:exp-main}
\end{table*}

\begin{figure*}
    \centering
    \includegraphics[width=1\linewidth]{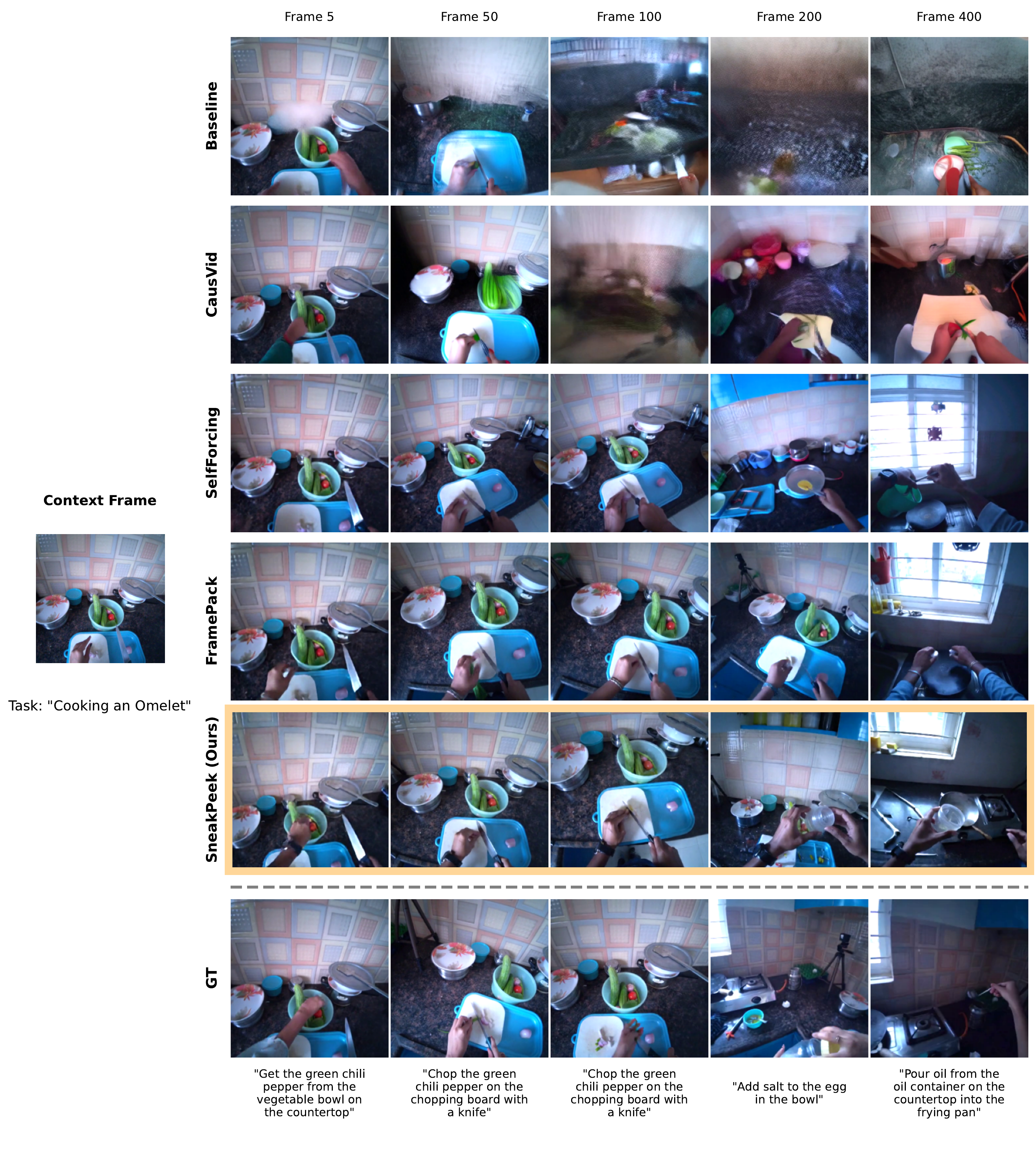}
    \caption{
        \textbf{Qualitative results comparing with prior methods on long video generation.}
        Compared to video frames generated by other methods in which suffer temporal drifting in long sequences or fail to follow step-wise prompts, that of ours are more temporally consistent and better prompt-aligned.
    }
    \label{fig:exp-qual}
\end{figure*}

\begin{figure*}
    \centering
    \includegraphics[width=1\linewidth]{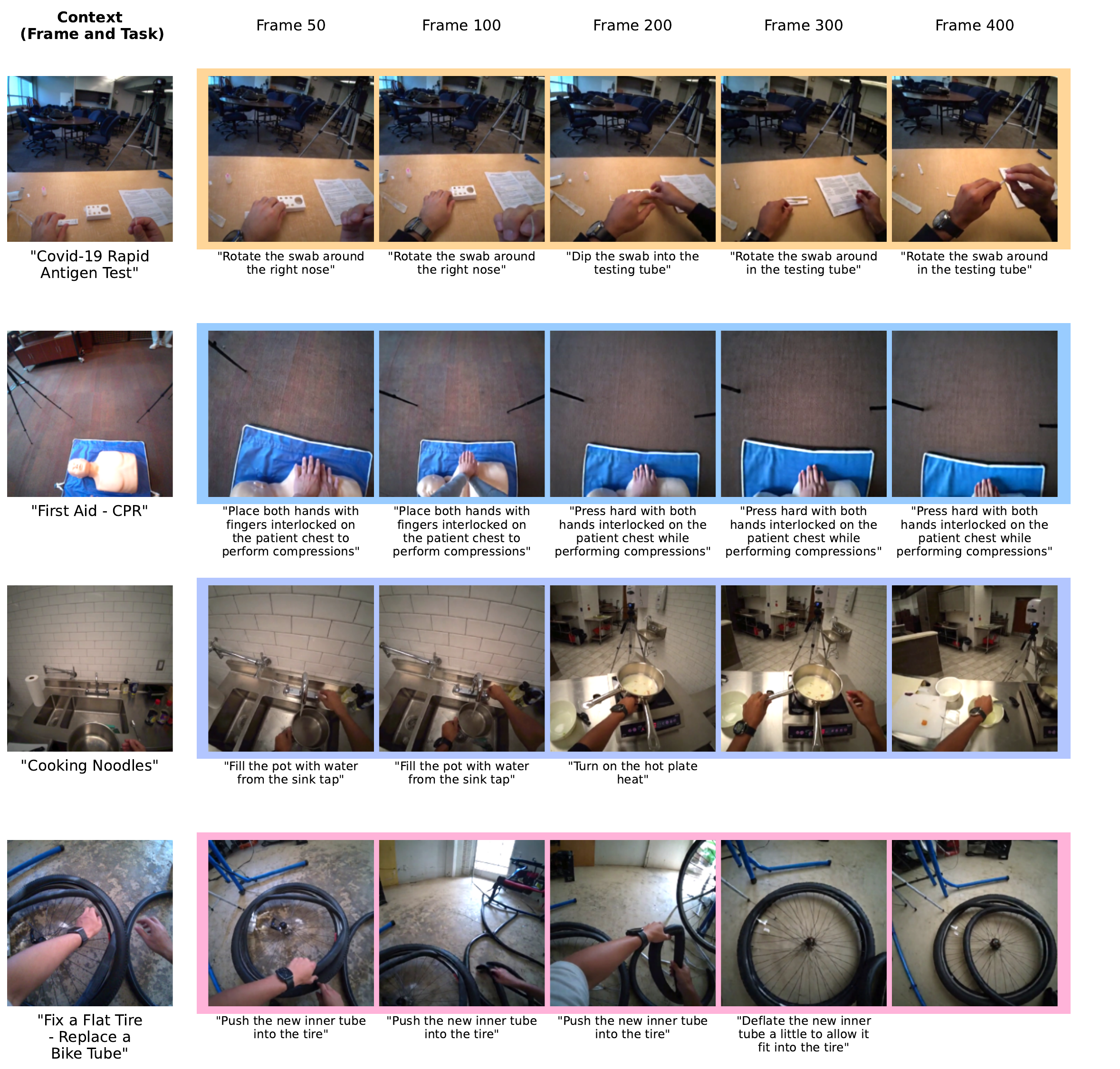}
    \caption{
        \textbf{Qualitative results of our method on different task domains.}
        Our framework demonstrates strong generalization across diverse instructional domains, showing consistent performance on tasks ranging from cooking and repair to medical procedures.
    }
    \label{fig:exp-qual-ours}
\end{figure*}

\paragraph{Dataset.}
We use the Ego-Exo4D-Keystep dataset~\cite{grauman2024ego}, which consists of multi-step instructional videos recorded from both egocentric and exocentric viewpoints. The dataset provides fine-grained temporal annotations for each step within a task (\eg, ``cut onions'' step from the ``cooking scrambled eggs'' task). We use the task name as the global prompt, while the local prompts correspond to the annotated step descriptions.

\paragraph{Data preparation for instructional generation.}
We use the egocentric recordings in Ego-Exo4D-Keystep~\cite{grauman2024ego} dataset, captured using Meta Aria smart glasses equipped with rolling-shutter RGB cameras. We use the center crop of the egocentric RGB videos, retaining the central 70\% of each frame to remove fisheye distortion. All videos are then resized to a spatial resolution of $480\times480$ and resampled to 16 FPS. Each long instructional video is divided into overlapping 1-minute clips to increase data diversity. The start time of each clip is aligned with the beginning of a key step annotation. Due to the dense annotations, each 1-minute clip contains on average 10 steps. In total, our processed dataset contains 11,905 1-minute clips (approximately 200 hours in total), covering three primary domains: 126 hours of cooking tasks, 20 hours of bike repair tasks, and 53 hours of health-related tasks (\eg, CPR, PCR, and COVID testing scenarios). We split this dataset into training and validation sets in a 9:1 ratio, with all clips derived from the same original video assigned to the same split to prevent data leakage. For the image-conditioned instructional video generation setting, the first frame of each clip serves as the conditioning image, providing the initial visual context for generation.

\paragraph{Evaluation metrics.}
We evaluate our model using a set of metrics to assess semantic consistency with textual instructions and visual-temporal coherence of the generated videos. To measure text-video consistency, we use the \textbf{ViCLIP}~\cite{wang2023internvid} score, which quantifies the alignment between the generated video and the global prompt (task). We further introduce \textbf{ViCLIP-M}, which computes the average ViCLIP similarity between step-wise segments and their corresponding local prompts, capturing fine-grained alignment at the procedural level. For image-video consistency, we employ two metrics: \textbf{ImageCLIP}~\cite{radford2021learning}, which measures whether the visual elements in the generated frames remain consistent with the initial conditioning image (\eg, utensils or background), and \textbf{ViCLIP-Image}~\cite{wang2023internvid}, which quantifies video-image semantic alignment over time. To assess temporal smoothness, we use \textbf{CLIP-F}~\cite{radford2021learning}, a frame-wise feature consistency metric reflecting motion stability and inter-frame coherence. Finally, we evaluate video-level similarity between the generated and ground-truth sequences using \textbf{EgoVLP}~\cite{lin2022egocentric}, a pretrained egocentric video-language model capturing both motion dynamics and action semantics. Together, these metrics comprehensively measure the fidelity, consistency, and procedural alignment of the generated instructional videos.

\subsection{Implementation details}

\paragraph{Training.}
For predictive causal adaptation, we use an internal pretrained bidirectional DiT model that operates in the latent space, as our base architecture. It is capable of generating 5-second clips. We adapt it to a 4-step model with causal attention mask and finetune using regression and cosine similarity losses. The training is performed for $8\times10^{3}$ iterations, using AdamW with a learning rate of $1\times10^{-5}$ and a batch size of 1. For each iteration, 21 latent frames are generated at once and random masking is applied per 3-latent-frame chunks.

During the future-guided self-forcing stage, we additionally finetune our 4-step model with flow-matching loss, and adversarial loss, where the discriminator operates on features extracted from the bidirectional base model. It is trained with AdamW optimizer, with a learning rate of $5\times10^{-5}$, $8\times10^{3}$ iterations, and a batch size of 1. We restrict our dual-region KV cache size to storing 9 latent frames for computational efficiency. The entire training process takes around two days using 16 GB200 GPUs.

\paragraph{Evaluation.}
During inference, we use 4 denoising steps to generate 30-second videos (480 video frames at 16 FPS) with a resolution of $480\times480$. Evaluation is performed on the Ego-Exo4D-KeyStep~\cite{grauman2024ego} validation set. The model synthesizes videos by iteratively generating 3-latent-frame chunks, each conditioned on the previously generated frames (\ie, roll-out). To improve temporal stability, every six rollout iterations, we set the model to predict future keyframes positioned 18 latent frames ahead, which are integrated into the attention cache through our dual-region KV caching mechanism. Then, the next frames are autoregressively generated until it reaches the predicted future KF, attending to both past and predicted future contexts.

\begin{table*}[t]
\renewcommand{\arraystretch}{1.1}
\centering
\aboverulesep=0ex
\belowrulesep=0ex
\caption{ 
    \textbf{Ablation study on each attribute:}  Predictive causal adaptation (PCA), future-guided self-forcing (F-SF), and temporal masking (TM) for fine-grained instruction control.
}
\resizebox{0.99\textwidth}{!}{
    \begin{tabular}{ccc|cc cc cc}
        \toprule
        \multirow{2}{*}{PCA} & \multirow{2}{*}{F-SF} & \multirow{2}{*}{TM} & \multicolumn{2}{c}{Text-Video Consistency} & \multicolumn{2}{c}{Image-Video Consistency} & Video Smoothness & Video Similarity \\
        \cmidrule(lr){4-5} \cmidrule(lr){6-7} \cmidrule(lr){8-8} \cmidrule(lr){9-9} 
         &&& ViCLIP$_\uparrow$ & ViCLIP-M$_\uparrow$ & ImageCLIP$_\uparrow$ & ViCLIP-I$_\uparrow$ & CLIP-F$_\uparrow$ & EgoVLP$_\uparrow$ \\
        \midrule
        \xm & \xm & \xm & 0.1358 & 0.1344 & 0.7896 & 0.7177 & \textbf{0.9848} & 0.4809\\
        \cm & \xm & \xm & 0.1472 & 0.1475 & 0.7867 & 0.7093 & 0.9680 & 0.4870\\
        \cm & \cm & \xm & 0.1602 & 0.1547 & 0.7701 & 0.6834 & 0.9685 & 0.4770\\
        \xm & \xm & \cm & 0.1515 & 0.1498 & 0.7903 & 0.6966 & 0.9606 & 0.4987 \\
        \cm & \xm & \cm & 0.1443 & 0.1345 & 0.7736 & 0.7044 & 0.9674 & 0.4804 \\
        \cm & \cm & \cm & \textbf{0.1663} & \textbf{0.1609} & \textbf{0.8153} & \textbf{0.7750} & \underline{0.9722} & \textbf{0.5445} \\
        \bottomrule
    \end{tabular}
}
\label{tab:exp-ablation}
\end{table*}

\subsection{Comparison with the state-of-the-art}

\paragraph{Compared methods.}
We compare our approach with three representative autoregressive video generation methods: Self-Forcing~\cite{huang2025self}, CausVid~\cite{yin2025slow}, and FramePack~\cite{zhang2025packing}. These methods are applied to the base bidirectional DiT model that represents high-quality video synthesis under full temporal context, denoted as Baseline. To evaluate the baseline model which is limited to generating short 5-second clips, we perform the rollout to longer sequences by using each clip's last frame as the context for the next. All methods are finetuned on the Ego-Exo4D-KeyStep~\cite{grauman2024ego} dataset using the same training and pre-processing setting as our model. For a fair comparison on instructional video generation, we apply our temporal masking strategy to all methods for multi-step instruction control.

\paragraph{Quantitative results.} 
\Cref{tab:exp-main} presents a quantitative comparison of our method against the Baseline, Self-Forcing~\cite{huang2025self}, CausVid~\cite{yin2025slow}, and FramePack~\cite{zhang2025framecontextpackingdrift} across text-video, image-video, and motion consistency metrics. Our approach achieves the highest scores in both text-video consistency metrics, particularly for ViCLIP-M, indicating superior alignment between the generated videos and multi-step textual instructions. Compared to the baseline, our method shows substantial improvements in EgoVLP and CLIP-F, demonstrating improved temporal coherence and reduced motion drift over long sequences.  While Self-Forcing and CausVid mitigate short-term inconsistencies, they exhibit gradual degradation over time, leading to blurred or frozen frames, as reflected in image-video consistency metrics. FramePack reduces temporal drift preserving image-video consistency, yet, its ability to follow step-wise instructions remains limited. Overall, these results confirm that our model effectively improves both temporal stability and instructional fidelity in long-horizon video generation.

\paragraph{Qualitative results.} 
We present qualitative comparisons with previous methods in~\Cref{fig:exp-qual}. While prior methods often suffer from temporal drift and inconsistent motion across long sequences, our method maintains smooth and coherent temporal transitions throughout the video. Compared to the Baseline and CausVid~\cite{yin2025slow}, which tends to freeze or collapse during long-horizon generation, our approach produces consistent motion trajectories guided by predicted future keyframes. Self-Forcing~\cite{huang2025self} and FramePack~\cite{zhang2025framecontextpackingdrift}  partially mitigate drift but still exhibit instability and limited capability of following step-wise instructions. In contrast, our method effectively preserves global temporal structure while enabling fine-grained control via multi-prompt conditioning. For example, generated results of ours show accurately transition from ``chopping the green chili pepper'' to ``adding salt'', while maintaining the same scene layout and dynamics. These qualitative results demonstrate that our approach achieves more controllable and motion-stable instructional video generation compared to existing methods.

Furthermore, we qualitatively evaluate our method across diverse instructional domains, including cooking, bike repair, and COVID test procedures. As shown in~\Cref{fig:exp-qual-ours}, our model generates coherent and contextually accurate videos across all categories. In cooking tasks, the model accurately follows sequential actions such as ``filling the pot with water" and ``turning on the heat", while preserving scene layout and object appearance. For bike repair, it captures fine-grained tool interactions and stable hand-object coordination over long sequences. In the COVID test and CPR scenario, our model successfully depicts subtle procedural transitions. These results highlight the strong generalization ability of our approach to long-horizon, multi-step instructional tasks. Additional qualitative results are provided in the supplementary materials.

\paragraph{Real-time generation.}
To evaluate the feasibility of streaming with our model, we built a distributed system that partitions the workload across multiple GPUs: four GPUs run the autoregressive generation, two GPUs predict future keyframes on demand, and one GPU decodes the latent video into RGB frames. With this setup on GB200 GPUs, our compiled model achieves 24 FPS at $480\times480$ resolution using FlashAttention 4~\cite{dao2022flashattention}, and delivers the first frame within 0.5 seconds.

\subsection{Ablation study}
We conduct an ablation study to analyze the contribution of each proposed component: Predictive Causal Adaptation (PCA), Future-guided Self-Forcing (F-SF), and Temporal Masking (TM) for fine-grained instruction control, as shown in \Cref{tab:exp-ablation}. The base model (first row of \Cref{tab:exp-ablation}) produces repetitive videos, obtaining high video smoothness score, yet significantly low text-video consistency scores. Adding PCA significantly increases both text-video and video-image alignment scores, indicating that the adaptation from the bidirectional teacher is crucial for learning temporally coherent motion under causal constraints. Incorporating F-SF further improves ViCLIP-M and EgoVLP metrics, suggesting that conditioning on predicted future keyframes enhances long-term consistency and reduces temporal drift. The TM mechanism provides additional gains in text-video alignment, especially for ViCLIP-M, demonstrating its importance for fine-grained control in multi-step procedural videos. The combination of all three components yields the best overall performance across metrics, achieving the highest text-video and image-video consistency while maintaining strong video smoothness.

\section{Conclusion}
\label{sec:conclusion}
In this work, we introduced a framework for instructional streaming video generation, addressing the challenges of long-horizon consistency and fine-grained procedural controllability. Our approach combines predictive causal adaptation, which enables robust future keyframe prediction under partial temporal contexts, with future-guided self-forcing that establishes bidirectional dependencies between current and predicted future frames to mitigate temporal drifting. Furthermore, temporal masking for multi-prompt conditioning provides step-specific controllability. Experimental results show that our method produces more stable motion, reduced temporal drift, and higher fidelity to multi-step textual instructions. We believe this work represents a step toward controllable, real-time procedural video generation capable of visualizing and updating instructional content on the fly.

\noindent\textbf{Limitations.}
Our method still accumulates errors across predicted keyframes, causing temporal drift over minute-long durations; future work will address this through Self-Forcing applied to keyframe prediction axis.

{
    \small
    \bibliographystyle{ieeenat_fullname}
    \bibliography{main}
}

\clearpage
\setcounter{page}{1}
\maketitlesupplementary

In this supplementary material, we present additional analysis of temporal drifting in \Cref{sup:temporal_drifting} and different sampling modes in \Cref{sup:sampling_modes}. Moreover, we validate the dynamic controllability of our framework in \Cref{sup:controllable_generation}. Also, we provide additional qualitative results in \Cref{sup:qual}.

\section{Additional analyses}
\label{sup:analysis}

\subsection{Temporal drifting}
\label{sup:temporal_drifting}
We evaluate temporal drift across methods by measuring the CLIP embedding distance between each generated frames and the initial reference frame. A larger value indicate greater deviation and thus weaker long-term stability. To ensure that drift measurements reflect temporal degradation rather than semantic changes caused by prompt transitions, we compute drift over a 10-second clip generated from a single instruction. To analyze how drift evolves over time, we report both the drift ratio (the linear slope of the drift curve) and drift acceleration (the quadratic coefficient indicating whether drift worsens progressively). Lower values in time-evolving metrics and absolute drift scores reflect more stable motion and reduced long-horizon degradation.

\begin{figure}
    \centering
    \includegraphics[width=1\linewidth]{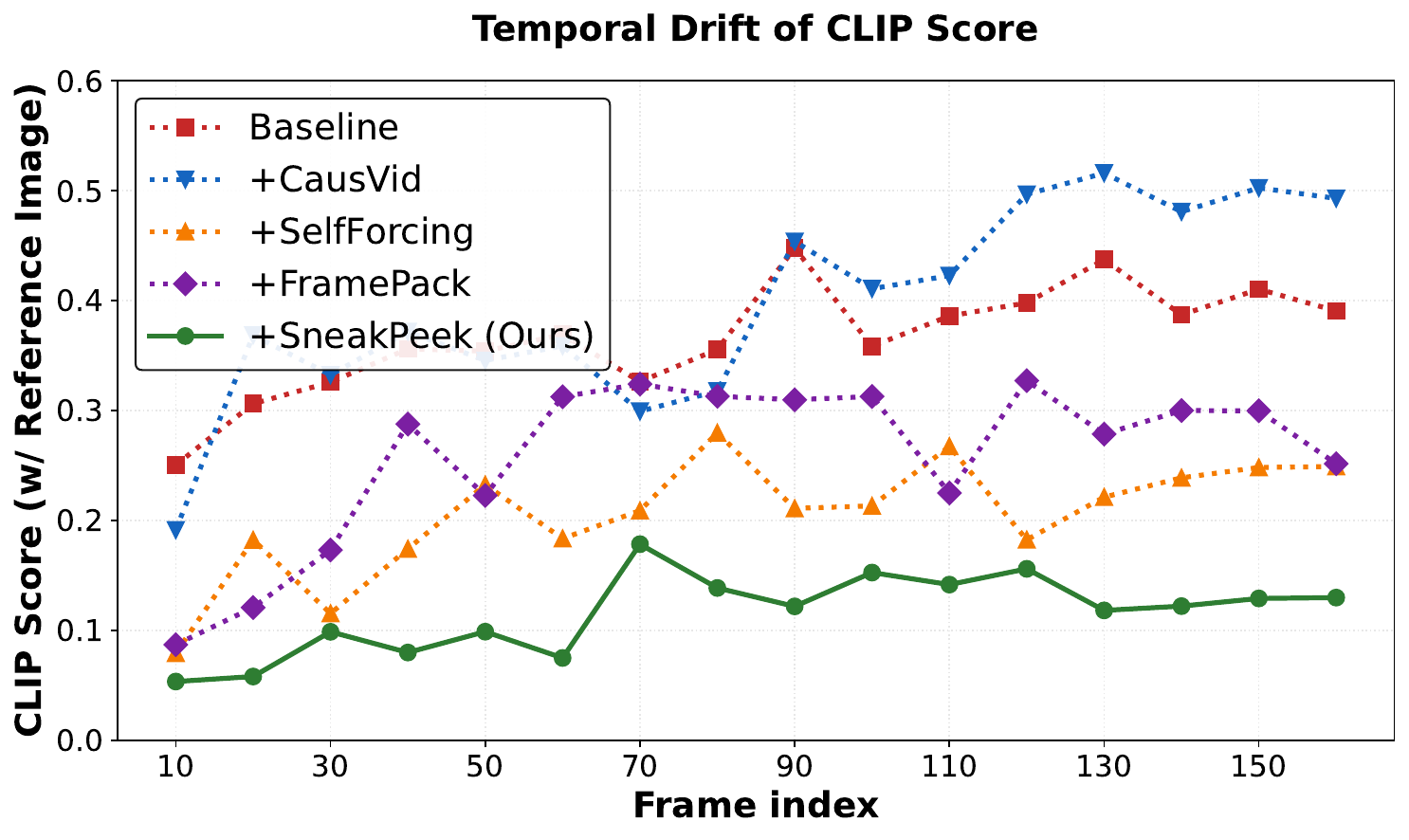}
    \caption{
        \textbf{Comparison of temporal drift trajectories.}   
    }
    \label{fig:sup-drift}
\end{figure}

As shown in~\Cref{tab:sup-drift}, our method achieves lowest overall drift (average and max) among all compared methods, while maintaining a negative drift acceleration. This indicates not only reduced drift severity but also that errors do not compound over time. In contrast, the baseline and CausVid~\cite{yin2025slow} exhibit higher drift average, max values, and drift ratio, indicating faster and less controlled drift accumulation. While SelfForcing~\cite{huang2025self} and FramePack~\cite{zhang2025framecontextpackingdrift} provide noticeable improvements, ours matches or surpasses them in stability and long-term robustness. Consistent with these trends, the visualization in~\Cref{fig:sup-drift} further shows that our approach maintains the most stable trajectory over extended durations. Such results confirm that our future-guided generation effectively suppresses temporal drift in videos.

\subsection{Sampling modes}
\label{sup:sampling_modes}
\begin{figure*}
    \centering
    \includegraphics[width=0.85\linewidth]{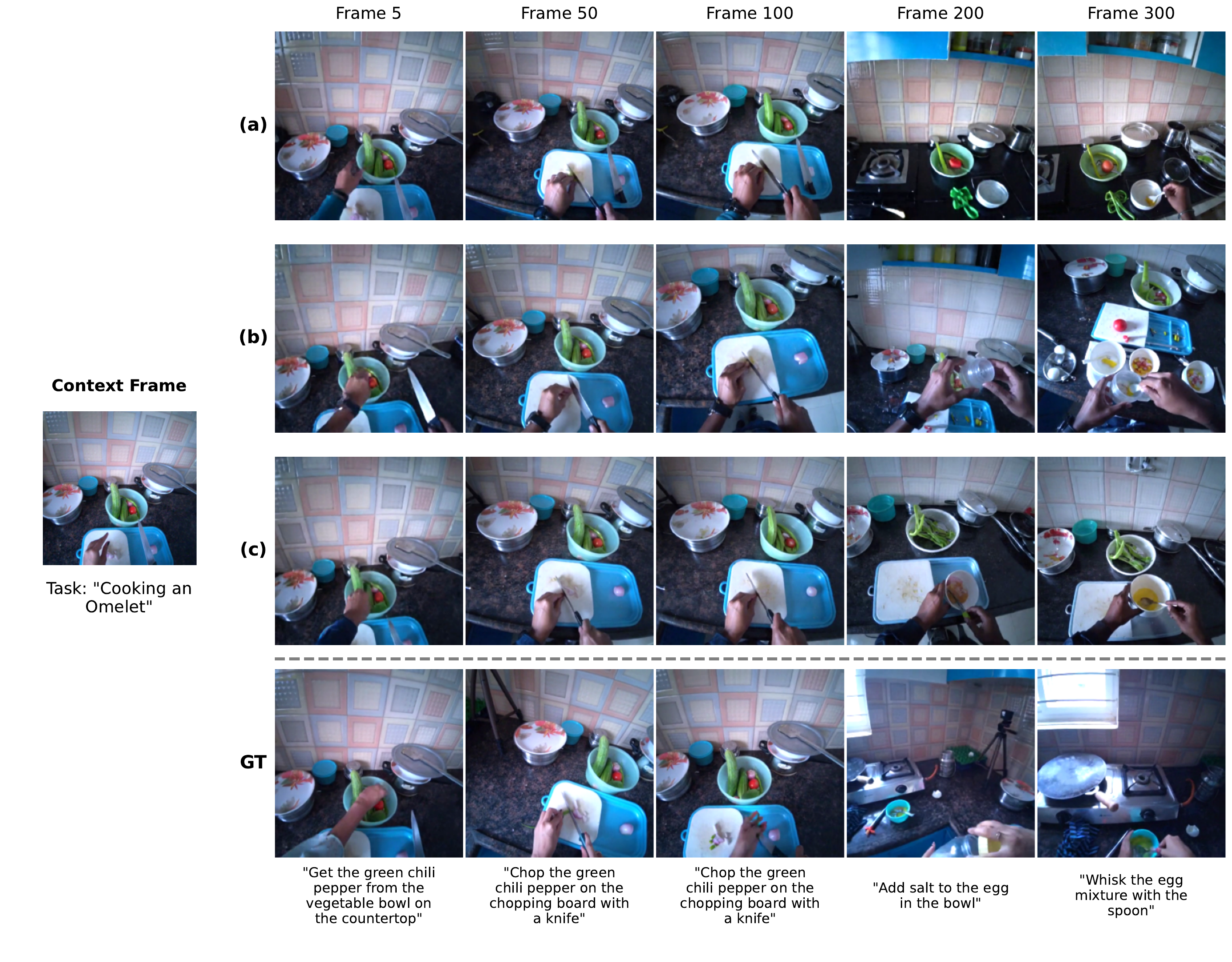}
    \vspace{-0.2cm}
    \caption{
        \textbf{Videos generated using different sampling modes}:
        (a) Full next-frame prediction,
        (b) Future KF prediction every 18 latent frames, where subsequent next frames are guided by the predicted future KF,
        (c) Future KF prediction every 9 latent frames, resulting in more frequent and short-ranged future guidance.
    }
    \label{fig:sup-sampling}
\end{figure*}

We analyze the behavior of our model under different sampling modes. Specifically, we compare 30-second videos generated by a sampling scheme: (1) fully consists of next-frame prediction without future keyframe (KF) predictions, (2) predicts future KF every 18 latent frames, and (3) every 9 latent frames. As shown in \Cref{fig:sup-sampling}, while the pure next-frame mode without future KF prediction (\Cref{fig:sup-sampling}a) exhibits noticeable degradation emerges after 200-th frame, incorporating periodic future KF prediction (\Cref{fig:sup-sampling}b) substantially mitigates long-term temporal drift. Increasing the KF prediction frequency (\Cref{fig:sup-sampling}c) suppresses drift, though it may introduce abrupt motion transitions or skipping intermediate instructional steps to satisfy the predicted future target within a shorter temporal window.

\subsection{Controllable generation}
\label{sup:controllable_generation}
\begin{figure*}
    \centering
    \includegraphics[width=0.85\linewidth]{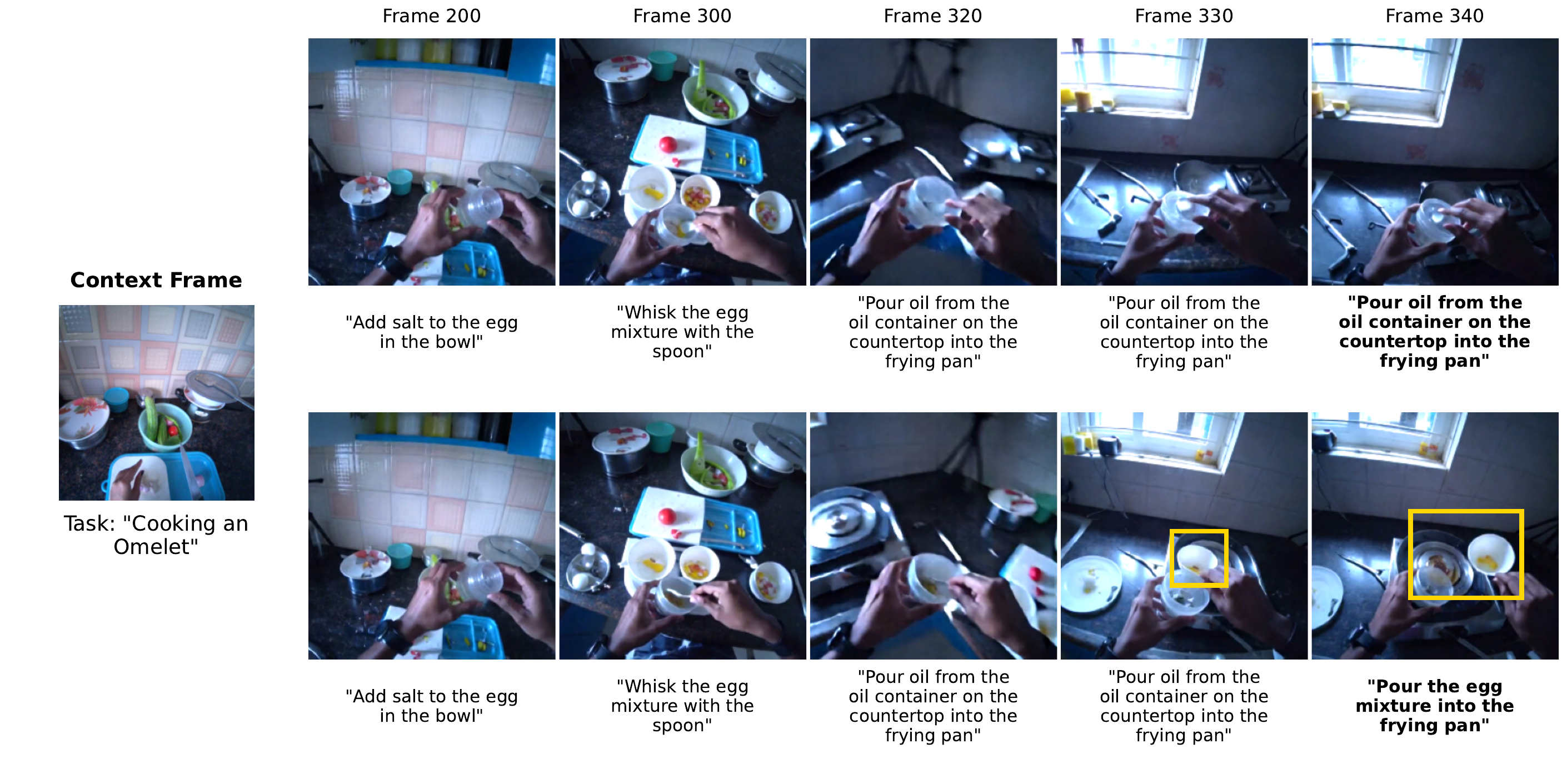}
    \vspace{-0.2cm}
    \caption{
        \textbf{Dynamic adaptability of our framework.}
        Modifying the conditioning prompt for the future KF changes the predicted KF, which in turn steers subsequent next frame predictions toward the newly anticipated future state.
    }
    \label{fig:sup-adaptability}
\end{figure*}

We examine our framework’s dynamic adaptability by modifying the prompt that conditions future KF prediction, as shown in \Cref{fig:sup-adaptability}. The model produces distinct future KFs for different prompts, and the intermediate frames adjust smoothly to these changes. The results demonstrate effective real-time responsiveness of our framework to evolving future instructions.

\begin{table}[t]
\renewcommand{\arraystretch}{1.1}
\centering
\aboverulesep=0ex
\belowrulesep=0ex
\caption{ 
    \textbf{Analysis of temporal drift.} Drift is measured using CLIP embeddings on every 10 frames of the generated video (16 FPS) using Ego-Exo4D-KeyStep validation set.
}
\resizebox{0.99\linewidth}{!}{
    \begin{tabular}{l|cccc}
        \toprule
        \multirow{2}{*}{Method} & \multicolumn{4}{c}{Drift measurement}\\
        \cmidrule(lr){2-5}
        & Average & Max & Ratio & Acceleration\\
        \midrule
        Baseline & 0.3610 & 0.4516 & 2.2733 & -1.217E-05\\
        + CausVid~\cite{yin2025slow} & 0.3473 & 0.4535 & 2.8801 & -9.787E-06 \\
        + SelfForcing~\cite{huang2025self} & 0.1799 & 0.2566 & 1.7795 & -3.998E-06\\
        + FramePack~\cite{zhang2025framecontextpackingdrift} & 0.1609 & 0.2370 & 1.6230 & -2.017E-06 \\
        + \Name (\textbf{Ours}) & 0.1489 & 0.2248 & 1.8035 & -5.767E-06 \\
        \bottomrule
    \end{tabular}
}
\label{tab:sup-drift}
\end{table}

\section{Additional qualitative results}
\label{sup:qual}
\begin{figure*}
    \centering
    \includegraphics[width=1\linewidth]{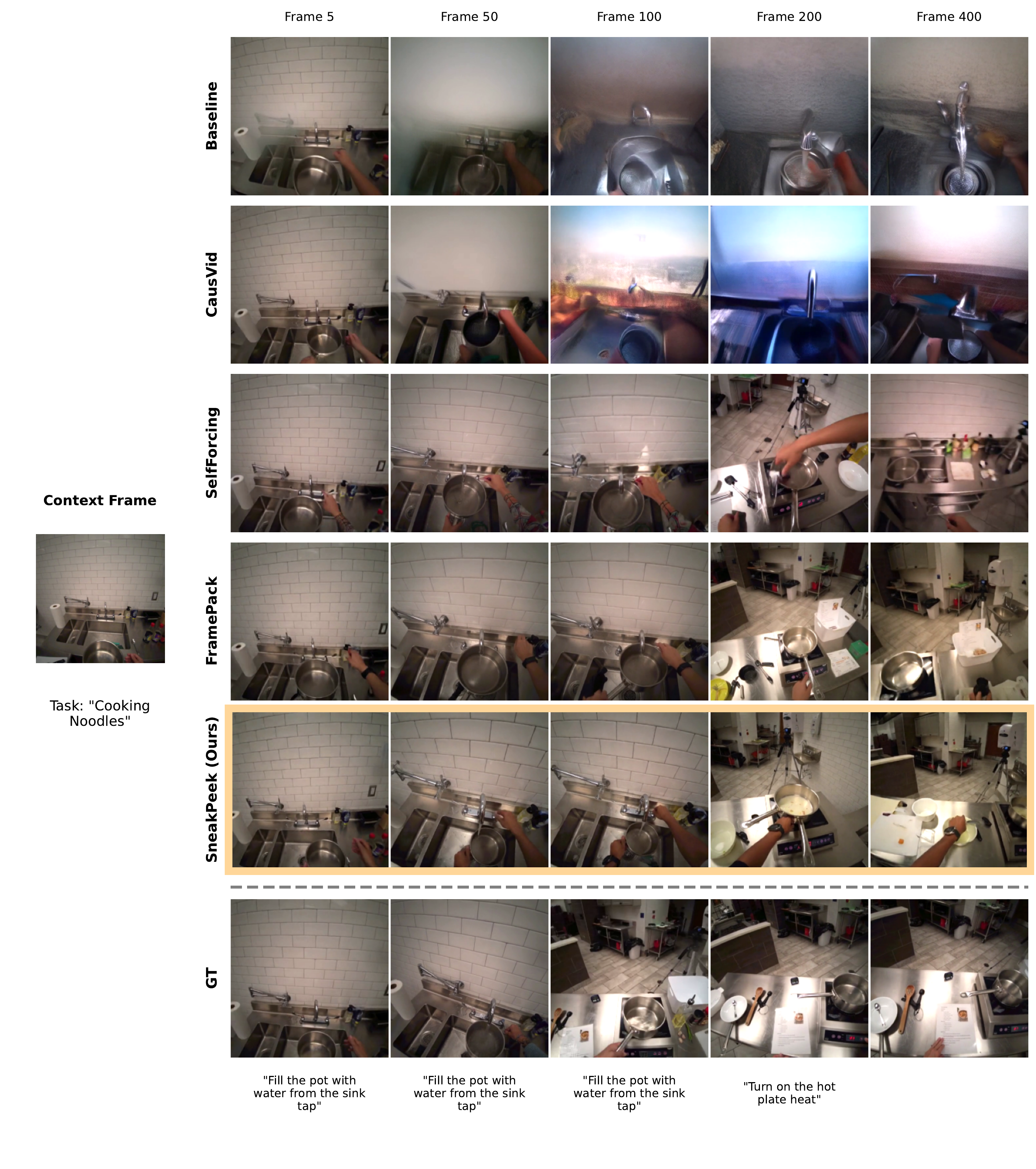}
    \caption{
        \textbf{Qualitative results on cooking task}. 
        Context frame and prompts are processed from `indiana\_cooking\_22\_5' of Ego-Exo4D-KeyStep.
    }
    \label{fig:sup-qual}
\end{figure*}

\begin{figure*}
    \centering
    \includegraphics[width=1\linewidth]{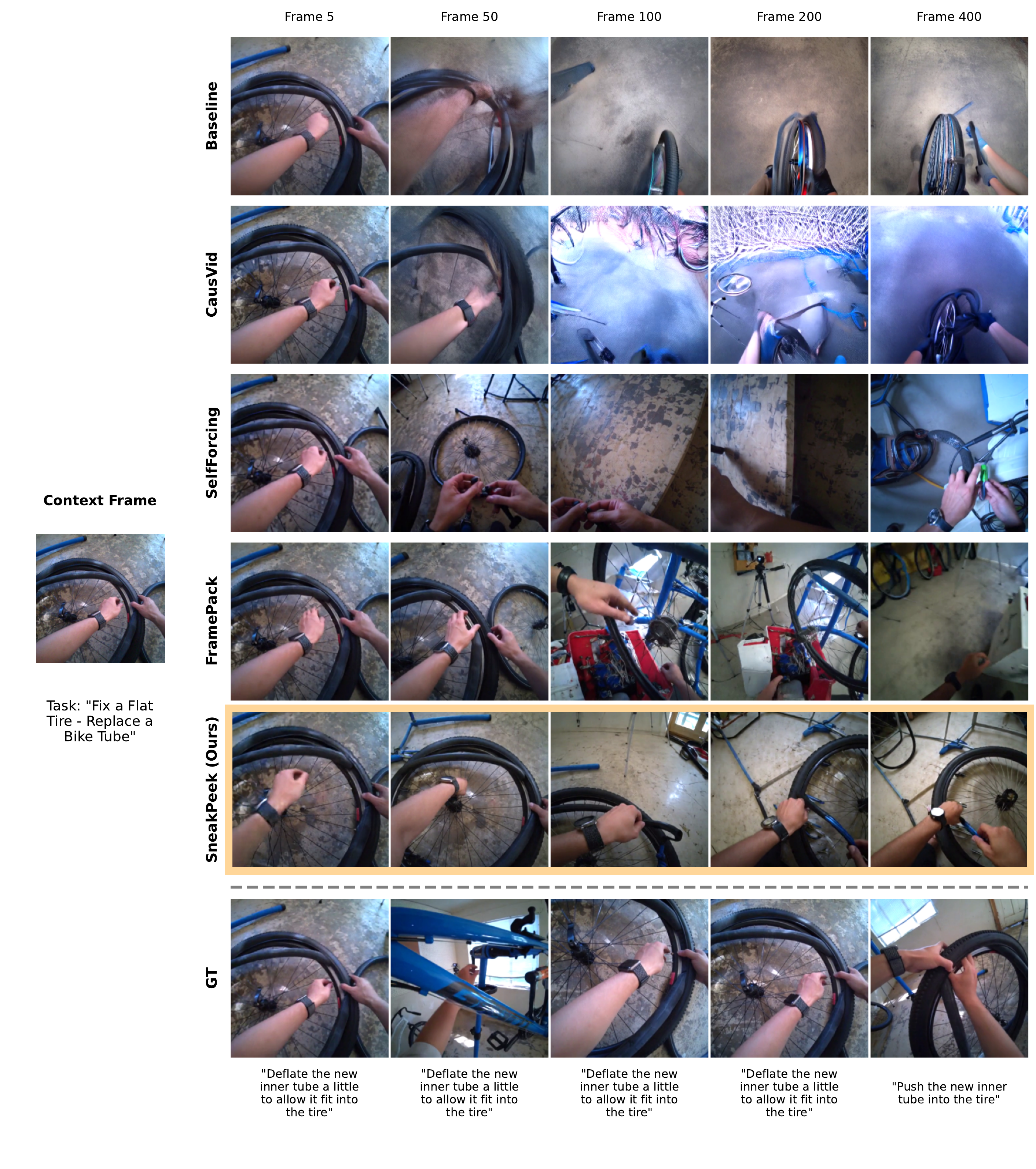}
    \caption{
        \textbf{Qualitative results on bike repair task}.
        Context frame and prompts are processed from `indiana\_bike\_12\_4' of Ego-Exo4D-KeyStep.
    }
    \label{fig:sup-qual2}
\end{figure*}

We provide additional visualization results to validate the generalizability of the findings in the main manuscript; compared to previous methods, our future-guided framework achieves lower temporal drifting and better step-specific controllability. According to~\Cref{fig:sup-qual} and~\Cref{fig:sup-qual2}, on different task domains, our method relieves temporal drifting over long sequences and achieves fine-grained control of segment prompts.


\end{document}